\UseRawInputEncoding
\documentclass[conference]{IEEEtran}
\IEEEoverridecommandlockouts
\usepackage{cite}
\usepackage{multirow}
\usepackage{verbatim}
\usepackage{amsmath,amssymb,amsfonts}
\usepackage{algorithmic}
\usepackage{graphicx}
\usepackage{textcomp}
\usepackage[dvipsnames]{xcolor}
\usepackage{url}
\usepackage{hyperref}
\usepackage[editing]{coop-writing}
\newcommand*{\crosssymbol}{%
    \text{%
      \raise 1ex\hbox{%
        \rlap{\vrule height.2pt depth.2pt width .75ex}%
        \hbox to .75ex{\hss\vrule height .5ex depth 1ex\hss}%
      }%
    }%
}

\usepackage{xparse}
\usepackage{listings}
\usepackage{xcolor}

\definecolor{eclipseStrings}{RGB}{42,0.0,255}
\definecolor{eclipseKeywords}{RGB}{127,0,85}
\colorlet{numb}{magenta!60!black}

\lstdefinelanguage{json}{
    basicstyle=\normalfont\ttfamily,
    commentstyle=\color{eclipseStrings}, 
    stringstyle=\color{purple}\ttfamily,
    numbers=left,
    numberstyle=\scriptsize,
    stepnumber=1,
    numbersep=8pt,
    showstringspaces=false,
    breaklines=true,
    frame=lines,
    string=[s]{"}{"},
    comment=[l]{:\ "},
    morecomment=[l]{:"},
    literate=
        *{0}{{{\color{blue}0}}}{1}
         {1}{{{\color{blue}1}}}{1}
         {2}{{{\color{blue}2}}}{1}
         {3}{{{\color{blue}3}}}{1}
         {4}{{{\color{blue}4}}}{1}
         {5}{{{\color{blue}5}}}{1}
         {6}{{{\color{blue}6}}}{1}
         {7}{{{\color{blue}7}}}{1}
         {8}{{{\color{blue}8}}}{1}
         {9}{{{\color{blue}9}}}{1}
}
\def\BibTeX{{\rm B\kern-.05em{\sc i\kern-.025em b}\kern-.08em
    T\kern-.1667em\lower.7ex\hbox{E}\kern-.125emX}}
\begin{document}

\title{Game-Agnostic Value Functions \\through Automatic JSON Feature Extraction}

\author{\IEEEauthorblockN{Dien Nguyen, Diego Perez-Liebana}

\IEEEauthorblockA{\textit{Queen Mary University of London - Game AI Group} \\ 
{d.l.nguyen, diego.perez}@qmul.ac.uk}
}

\maketitle
\IEEEpubidadjcol

\begin{abstract}
JSON Bag-of-Tokens (JSON-Bag) is a recently proposed method to generically represent game trajectories by tokenizing their JSON descriptions. We introduce JSON-Bag VF, a game-agnostic approach to training value functions for game-playing agents using JSON-Bag prototypes. We show that this approach can be enhanced with Random Forest-based feature selection and a method to select game-stage-specific features. We evaluate JSON-Bag VF with One-step-look-ahead (JSON-Bag OSLA) on six tabletop games over different combinations of prototype-tokenization and feature selections. JSON-Bag OSLA outperforms baseline OSLA agents in most games. Our analysis also shows that feature selection significantly improves JSON-Bag VF and that feature selection is the most important factor in JSON-Bag VF performance, over prototype-tokenization.
\end{abstract}


\newcommand\width{0.75}
\section{Introduction}
Defining features and representations for games and their corresponding distance/similarity metric is foundational for any task that requires game analysis. Designing strong game-playing agents is such a task, often requiring hand-crafted features using domain knowledge \cite{silver_temporal-difference_2012} or expensive deep-learning models \cite{silver_mastering_2017}.

JSON Bag-of-Tokens model (JSON-Bag) \cite{nguyen_json-bag_2025} is a framework to generically represent game trajectories by tokenizing their JSON descriptions. The JSON representation for a game trajectory is formed by concatenating the JSON of individual game states, serialized from the game objects' data, into a list. A JSON-Bag is a collection of token-occurrence count pairs from tokenizing the game trajectory in JSON. JSON-Bag has only been tested on toy problems to classify game trajectories.

This work aims to close the gap between toy problems and practical application by using a game-agnostic method to train a value function (JSON-Bag VF) with JSON-Bag prototypes formed by a classification task of predicting the winning player, utilizing Jensen-Shannon distance (JSD) \cite{lin_divergence_1991} and prototype nearest-neighbor search (P-NNS) \cite{nguyen_json-bag_2025}. We demonstrate JSON-Bag's utility as an automatic feature extraction tool through Random Forest-based feature selection, improving JSON-Bag VF performance over no feature selection. Additionally, we introduce a method for selecting game-stage-specific features by segmenting game trajectories and applying feature selection to each segment independently.

We validate our approach on six tabletop games: \textit{7 Wonders}, \textit{Dominion}, \textit{Sea Salt and Paper}, \textit{Can't Stop}, \textit{Connect4}, and \textit{Dots and Boxes}. For each game, we evaluate different prototype-tokenization strategies and feature selection methods. We use One-step-look-ahead (OSLA) with trained JSON-Bag VFs (JSON-Bag OSLA) as our proof-of-concept agent, with more sophisticated agents such as Monte Carlo Tree Search (MCTS) \cite{browne_survey_2012} left for future work. JSON-Bag OSLA significantly outperforms standard OSLA agents, which use the game scoring as value functions, across all games, except \textit{Sea Salt and Paper} and \textit{Dots and Boxes}. We analyze the effect sizes of prototype-tokenization and feature selection, showing that the latter is the most important factor for the agent's performance.

The main contributions of this paper are:
\begin{itemize}
  \item \textbf{JSON-Bag VF}: A game-agnostic approach for training a value function using JSON-Bag prototypes from a winner prediction task, leveraging JSD and P-NNS.
  \item We show JSON-Bag compatibility with classical feature selection methods by greatly improving the performance of the value function with Random Forest-based feature selection. We also describe a method for selecting game-stage-specific features by segmenting game trajectories and applying feature selection separately to each segment.
  \item Through effect size analysis, we show that feature selection is the most important factor for the agent's score.
\end{itemize}


\section{Games}
The following $6$ games are used in our experiments, using their implementation in the TAG framework \cite{gaina_tag_2020}.

\textbf{7 Wonders} (Antoine Bauza, 2010): Players draft cards to build their civilizations, interacting with neighbors by passing cards and buying resources for their unique assigned wonder. 

\textbf{Dominion} (Donald X. Vaccarino, 2008): A deck-builder game where players create an ``engine" to acquire victory point cards in the late game. We use the ``First Game" card setup. 

\textbf{Sea Salt and Paper} (Bruno Cathala, 2022): A set-collection game where players draw cards to create card combos that maximize their score. Players can choose when and how to end the round---higher risk options can gain or lose them points.

\textbf{Connect4} (Milton Bradley, 1974): Players take turns dropping board pieces into a vertical grid to connect four of their pieces in a row, column, or diagonal before their opponent. 

\textbf{Dots and boxes} (Edouard Lucas, 1889): Players take turns placing a link between adjacent dots on a grid, until filled. If a player forms a box, they score one point and play again.

\textbf{Can't Stop} (Sid Sackson, 1980): A push-your-luck dice game with eleven number tracks, from 2 to 12. Players roll four dice to form two sums and advance markers on corresponding number tracks, aiming to complete three tracks before their opponents. Players must decide whether to stop or keep rolling to advance and risk losing progress.

\section{JSON Bag-of-tokens model}
The JSON Bag-of-tokens model (JSON-Bag) represents game trajectories by tokenizing their JSON representation.
The JSON of a game state is made by serializing the data of every game component, with the specific serialization structure depending on the game implementation. This JSON can be tokenized: a token is each of its individual \textit{atomic} components (non-collections), identified by a string containing the path from the outermost level to that component, each level separated by a dot. At each level, if the current element is in a list and the tokenization is \textit{ordered}, its index can also be encoded. For example, a snippet from a \textit{7 Wonders} JSON:
\begin{lstlisting}[language=json,numbers=none,basicstyle=\small\ttfamily]
{"currentAge": 2,
  "playerResources": [
    {"Wood": 2},
    {"Wood": 2}
]}
\end{lstlisting}
They can be tokenized both ordered and unordered as:
\begin{lstlisting}[language=json,numbers=none,basicstyle=\small\ttfamily]
".currentAge.2",
".playerResources.Wood.2",
".playerResources.Wood.2",
".playerResources[0].Wood.2",
".playerResources[1].Wood.2"
\end{lstlisting}

All game states of a game trajectory are tokenized and processed into a JSON-Bag, where the occurrences of each token are counted and combined into a token-occurrence count pair. Each game trajectory is a bag-of-tokens, and the occurrence values are normalized to sum to $1$ within a bag to model a probability distribution.
In this paper, we introduce a small enhancement to tokenization where a token ending in a positive integer $n$ of the form \lstinline[basicstyle=\small\ttfamily]{"X.Y.Z.n"}, will also be treated as $n$ separate instances of the token \lstinline[basicstyle=\small\ttfamily]{"X.Y.Z"}. All JSON-Bags in this paper are tokenized both \textit{ordered} and \textit{unordered}.


A \textbf{JSON-Bag prototype} is simply the average of all bags of that specific class in the dataset, forming a single JSON-Bag that represents an entire class of game trajectories (e.g., all game trajectories where player 1 wins the game). This is useful for distance-based classification.

Jensen-Shannon (JS) divergence \cite{lin_divergence_1991} is an information-theoretic divergence based on Shannon entropy to measure the similarity between distributions. The square-root of this gives \textbf{Jensen-Shannon Distance} (JSD), which is a \textit{metric} \cite{endres_new_2003} (i.e., satisfies triangle inequality).
For discrete distributions $P$, $Q$ defined on the same sample space and mixture distribution $M = (P+Q)/2$, the JSD between $P$ and $Q$ is:
\begin{equation}
    \label{jsd}
    Dist_{JS}(P, Q) = \sqrt{D_{KL}(P||M)/2 + D_{KL}(Q||M)/2}
\end{equation}
where $D_{KL}(P||Q)$ is the KL divergence between $P$ and $Q$. 

\textbf{Prototype Nearest-Neighbor Search} (P-NNS) classifies a JSON-Bag to the label of the closest prototype. JSON-Bags can be interpreted as distributions, thus using JSD as a distance metric is theoretically grounded and shown to perform better than Euclidean and Cosine distance in classification tasks \cite{nguyen_json-bag_2025}.

\section{JSON-Bag as Value Function (JSON-Bag VF)}
\textbf{Classifying the winning player} We formulate a classification task of predicting the winning player of a game trajectory. Given training data of full game trajectories represented as JSON-Bags, each trajectory is labeled with the winning player, then the \textit{prototype} of each label is the sum of all JSON-Bags with that label. The prototype of player $P_i$ (where $i$ is the player index) would represent the most likely distribution of tokens for a game trajectory in which $P_i$ is the winning player. Using P-NNS, the value of a game trajectory (from the perspective of $P_i$) could simply be a binary random variable in which it is $1$ if the predicted winning player is $P_i$, and $0$ otherwise. However, we can directly use the distances computed in P-NNS to derive a real-valued score to also indicate the confidence of the prediction.

Given a JSON-Bag game trajectory $\tau$, $d(\tau, i)$ is the JSD between $\tau$ and the prototype of $P_i$, the value function $V_i(\tau)$ from the perspective of $P_i$ is defined as:
\begin{equation}
    \label{val}
    V_i(\tau) = \frac{1-d(\tau, i)}{\sum_n{1-d(\tau, j)}}
\end{equation}
where $n$ is the number of players. This makes sure a smaller relative distance to prototype $P_i$ gives a larger value, implying that the trajectory is more likely to represent a winning trajectory for $P_i$. Note both JSD and $V_i$ are bounded between in $[0,1]$. We use a simple linear normalization to compute $V_i$ from a set of distances, although other methods are possible.

\textbf{JSON-Bag One-step-look-ahead (JSON-Bag OSLA)} JSON-Bag VF can be used with any decision-making algorithm; here, we test it with One-step-look-ahead (OSLA). The agent keeps track of the current game trajectory represented as a JSON-Bag, updated at the beginning of the agent's turn. This JSON-Bag can be interpreted as the agent's summary of the game trajectory thus far. During action evaluation, for each action $a$, the agent uses $a$ to roll the game state forward, then uses the new game state to get a new game trajectory $\tau_a$. The action value for $a$ is $V_i(\tau_a)$ where $i$ is the player index of the OSLA agent; the agent then deterministically chooses the action with the highest value. For imperfect information games, hidden information is redeterminized \cite{cowling_information_2012} before the game state is used by JSON-Bag OSLA.

\textbf{Feature Selection} JSON-Bag VF is formulated as a classification task. This lets us improve its performance with feature selection--only consider the important tokens and reduce noise from redundant ones. We apply \textbf{Random Forests} (RF) \cite{breiman_random_2001} to classify the winning player, treating each token as a feature and its frequency as the value to be used in RF. Using the resulting model, the top-$k$ features can be selected by ranking the features by their Mean Decrease in Impurity (MDI) \cite{breiman_random_2001}. 

For many games, the most important features are different depending on the stage of the game. We divide the full game trajectories into $N_S$ equal segments, then create $N_S$ datasets, where dataset $D_i$ has game trajectories up to the $i$-th segment. For example, the dataset $D_3$ would include game trajectories from the beginning of the game up to (and including) segment $3$. The feature selection procedure is applied to each dataset, combining the resulting top-$k$ features from each dataset into a final list of features, which can be larger than $k$. Game-specific knowledge could customize the segmentation to each game, potentially yielding better results.

\section{Experiments}
For each game, JSON-Bag OSLA is tested with different ways to tokenize winning player prototypes and select features.
The training data for each game is generated by $1000$ games of self-play using MCTS agents tuned for that game with an action-decision budget of $128$ms. The data includes the JSON description of every game state after a player ends their turn. 

When the training data is tokenized to turn into \textit{prototypes}, how often a game state is selected to be tokenized can vary. Full game trajectories can be tokenized in full, with every game being state tokenized. However, this risks oversampling certain tokens and potentially leading to noisier prototypes; we investigate if sparser sampling may help with this issue. We propose the following tokenization frequency strategies: 
\begin{itemize}
  \item \textbf{Full}: All game states from the training data are tokenized. 
  \item \textbf{Interval}: Game states are tokenized after every $f$ turns. Here, we choose $f = n$, the number of players. The terminal game state is always included.
  \item \textbf{Random}: Game states are tokenized with probability $p$. We choose $p = 1/n$, allowing sampled states from multiple players' turns, while a fixed-interval could lead to the tokenization of game states of only a single player.
\end{itemize}

We also test feature selection with $k=100$, using segmented ($N_S=4$) and non-segmented ($N_S=1$) game trajectories.
For each variant, we use the $3$ tokenization frequencies described above to build the JSON-Bags for Random Forests, resulting in $6$ variants of feature selection. 


Each game is tested with all pairs of tokenization frequencies and feature selection, employing a single JSON-Bag OSLA against baseline OSLA agents, for $1000$ games for each JSON-Bag OSLA player position. The baseline OSLA agents use the scoring of the game as their value function (e.g., Victory Points in Dominion). The \textit{score} of each game is -1, 0, 1 for a loss, draw, win for JSON-Bag OSLA, respectively. 


\section{Results and Discussion}
\begin{table*}[t]
	\centering
    \caption{Top-3 prototypes and feature pairs. Prototype-T and Feature-T are the tokenization frequencies used when making prototypes and selecting features, respectively. Score is mean $\pm$ standard error. Best-Worst Diff is the score difference between best and worst player position for each prototypes/features pair. Smaller values mean more similar scores across player positions.}
	\label{tab:overall-results}
    { 
	\begin{tabular}{l|c|c@{\hskip 4pt}|c@{\hskip 4pt}|c|c@{\hskip 4pt}|c@{\hskip 4pt}}
		\hline
		Game & n-players & Prototype-T & Feature-T & Segmented & Score & Best-Worst Diff \\
		\hline
		Dominion & 4 & all & interval & True & $0.981 \pm 0.003$ & 0.032 \\
		 &  & interval & interval & True & $0.980 \pm 0.003$ & 0.037 \\
		 &  & random & interval & True & $0.974 \pm 0.004$ & 0.051 \\
		\hline
		7 Wonders & 4 & random & all & False & $0.657 \pm 0.012$ & 0.037 \\
		 &  & interval & all & False & $0.656 \pm 0.012$ & 0.112 \\
		 &  & interval & interval & False & $0.638 \pm 0.012$ & 0.152 \\
		\hline
		Connect-4 & 2 & random & interval & False & $0.634 \pm 0.017$ & 0.308 \\
		 &  & interval & interval & False & $0.613 \pm 0.018$ & 0.286 \\
		 &  & all & interval & False & $0.611 \pm 0.018$ & 0.406 \\
		\hline
		Can't Stop & 4 & interval & interval & True & $0.595 \pm 0.013$ & 0.618 \\
		 &  & interval & interval & False & $0.594 \pm 0.013$ & 0.666 \\
		 &  & interval & all & False & $0.186 \pm 0.016$ & 1.920 \\
		\hline
		Dots And Boxes & 2 & random & all & False & $-0.451 \pm 0.020$ & 0.130 \\
		 &  & all & all & False & $-0.514 \pm 0.019$ & 0.524 \\
		 &  & all & random & False & $-0.557 \pm 0.019$ & 0.182 \\
		\hline
		Sea Salt and Paper & 4 & interval & random & True & $-0.717 \pm 0.011$ & 0.526 \\
		 &  & interval & all & True & $-0.753 \pm 0.010$ & 0.399 \\
		 &  & interval & interval & True & $-0.761 \pm 0.010$ & 0.393 \\
		\hline
	\end{tabular}
	}
	\centering
\end{table*}

\begin{table}[t]
	\centering
    \caption{Best performing prototypes with no feature selection.}
	\label{tab:no-fs-results}
	\begin{tabular}{c|c|c|c}
		\hline
		Game  & Prototype-T & Score & Best-Worst Diff \\
		\hline
		Dominion & interval & $0.233 \pm 0.015$ & 1.011 \\
		\hline
		7 Wonders & full & $0.134 \pm 0.015$ & 0.510 \\
		\hline
		Connect-4 & random & $0.530 \pm 0.019$ & 0.500 \\
		\hline
		Can't Stop & interval & $0.314 \pm 0.015$ & 0.512 \\
		\hline
		Dots And Boxes & random & $-0.868 \pm 0.011$ & 0.104 \\
		\hline
		Sea Salt and Paper & interval & $-0.796 \pm 0.001$ & 0.193 \\
		\hline
	\end{tabular}
\end{table}

\autoref{tab:overall-results} shows the result of top-3 prototypes/feature-selection pairs for each game, and \autoref{tab:no-fs-results} shows the result of the best performing prototype-tokenization for each game \textit{without} feature selection. "Best-Worst Diff" is the score difference between the best and worst performing player position for each prototypes/features pair, showing how varied the score is when an agent is played at different player positions. Smaller values indicate more similar scores across player positions. 

For \textit{Dominion}, \textit{7 Wonders}, \textit{Connect-4}, and \textit{Can't Stop}, JSON-Bag OSLA significantly outperforms baseline OSLA with or without feature selection, indicating JSON-Bag is encoding useful game features beyond just game scoring. In \textit{Dots and boxes} and \textit{Sea Salt and Paper}, JSON-Bag OSLA is noticeably worse regardless of feature selection. We hypothesize that this performance gap is due to the fact that game scores alone for these two games are already informative enough to have a strong agent. With JSON-Bag, the scores are tokenized but other game tokens are comparatively more irrelevant, leading to noisier JSON-Bag prototypes. In \textit{Sea Salt and Paper} specifically, the scores of players' current hands are not explicitly stored in the game state and have to be inferred from player hands, which JSON-Bag OSLA could not do since that requires more complex interaction between tokens.

In most cases, feature selection noticeably improves JSON-Bag OSLA performance, especially for \textit{Dominion} and \textit{7 Wonders}. \textit{Dominion} benefits from having a segmented feature selection, which makes sense given that, in the early-game, players are less focused on buying victory points than on generating money to later buy them. It is surprising, however, that \textit{7 Wonders}, a game designed with explicit game stages, benefits from non-segmented feature selection.

In some games, JSON-Bag OSLA observes large differences between the scores of its best and worst performing player positions, indicating a large performance gap depending on where JSON-Bag OSLA starts in the player order. For certain games and prototypes/features pairs, the gap is larger than what would be normally observed in baseline OSLA agents. This warrants further investigation in future work.

\textbf{Effect sizes of prototype-tokenization vs. feature selection}
\begin{table}[t]
	\centering
    \caption{Effect sizes ($\eta^2$) of categorical parameters on JSON-Bag OSLA performance across games. Larger $\eta^2$ value means greater contribution to explaining the agent's score. The parameters \textit{prototypes} and \textit{features} have 3 and 6 categories, respectively.}
	\label{tab:effect-sizes}
	\begin{tabular}{l|c|c}
		\hline
		Game & Parameter & $\eta^2$ \\
		\hline
		Dominion & prototypes & 0.0001\\
		 & features & 0.9992\\
		\hline
        7 Wonders & prototypes & 0.0019\\
		 & features & 0.9137\\
		\hline
        Connect-4 & prototypes & 0.0420\\
		 & features & 0.9070\\
		\hline
        Can't Stop & prototypes & 0.4409\\
		 & features & 0.4772\\
		\hline
		Dots And Boxes & prototypes & 0.1970\\
		 & features & 0.7211\\
		\hline
		Sea Salt and Paper & prototypes & 0.0872\\
		 & features & 0.8736\\
		\hline
	\end{tabular}
	\centering
\end{table}
\autoref{tab:effect-sizes} shows the effect sizes of tokenization frequency for prototypes vs. different feature selection variants, or how much varying prototype-tokenization and feature selection affects JSON-Bag OSLA's scores. The values of $\eta^2$, $0 \leq \eta^2 \leq 1$, indicate the proportion of variance in the agent's score that can be explained by the choice of feature selection, or how prototypes are tokenized, respectively. For most games, how features are selected is by far the most important factor in the agent's score, while \textit{Can't Stop} observes equal contribution between both. This further proves the importance of feature selection in improving performance of JSON-Bag VF.

\section{Conclusion}
We propose JSON-Bag VF, a game-agnostic approach to train a value function for game-playing agents using JSON-Bag prototypes, which are formed by a winner prediction task. JSON-Bag VF utilizes Jensen-Shannon distance (JSD) and prototype nearest-neighbor search (P-NNS) to derive the value of a JSON-Bag game trajectory from its distances to other JSON-Bag prototypes, each representing a different winning player. Feature selection using Random Forest is used to further improve the performance of JSON-Bag VF. We also introduce a method to select game-stage-specific features by segmenting individual game trajectories into multiple datasets and applying feature selection on each independently.

JSON-Bag VF is evaluated with a One-step-look-ahead (OSLA) agent against baseline OSLA agents over six tabletop games--\textit{7 Wonders}, \textit{Dominion}, \textit{Can't Stop}, \textit{Connect4}, \textit{Sea Salt and Paper}, \textit{Dots and boxes}, outperforming the baselines on all except the last two.
We also show that feature selection improves JSON-Bag VF performance on all games and, through effect size analysis, that feature selection is more important to an agent's score than how the 
prototypes are tokenized.

However, there are limitations to be addressed in future work. Firstly, JSON-Bag VF is only tested with simple OSLA agents. The results are promising, which motivates further investigating its use with more sophisticated agents.
Secondly, the choice of top-$k$ feature selection is arbitrary, given each game has a different number of total tokens. More principled approaches, such as choosing $k$ based on a threshold of MDI scores, might yield better results.
Finally, we've observed that the performance of the JSON-Bag value functions noticeably decreases when game scoring alone yields strong agents, or when the scoring is not clearly described in the game state but has to be inferred instead from other game elements. More sophisticated tokenization approaches, aimed to emphasize interaction between different tokens, 
could help with this issue. 


\section*{Acknowledgments}
For the purpose of open access, the author(s) has applied a Creative Commons Attribution (CC BY) license to any Accepted Manuscript version arising. This work was supported by the EPSRC Centre for Doctoral Training in Intelligent Games \& Games Intelligence (IGGI) EP/S022325/1.

\bibliographystyle{IEEEtran}
\bibliography{references}
\vspace{12pt}

\end{document}